\pdfoutput=1
\documentclass[11pt]{article}

\usepackage[]{acl}

\usepackage{times}
\usepackage{latexsym}

\usepackage[T1]{fontenc}
\usepackage[utf8]{inputenc}

\usepackage{microtype}

\usepackage{inconsolata}

\usepackage{amsthm,amsmath,amssymb}
\usepackage{textcomp}
\usepackage{multicol}
\usepackage{multirow}
\usepackage{graphicx}
\usepackage{bm}
\usepackage{booktabs}       % professional-quality tables
\usepackage{diagbox}
\usepackage{pifont}
\usepackage{float}

\usepackage{mdframed}
\usepackage{listings}

\title{Understanding and Addressing the Under-Translation Problem from the Perspective of Decoding Objective}

\author{
  Chenze Shao,~
  Fandong Meng,~ 
  Jiali Zeng,~
  and Jie Zhou \\
  Pattern Recognition Center, WeChat AI, Tencent Inc, China \\
  \texttt{\{chenzeshao, fandongmeng, lemonzeng, withtomzhou\}@tencent.com} \\
}

\begin{document}
\maketitle
\begin{abstract}
Neural Machine Translation (NMT) has made remarkable progress over the past years. However, under-translation and over-translation remain two challenging problems in state-of-the-art NMT systems. In this work, we conduct an in-depth analysis on the underlying cause of under-translation in NMT, providing an explanation from the perspective of decoding objective. 
To optimize the beam search objective, the model tends to overlook words it is less confident about, leading to the under-translation phenomenon. Correspondingly, the model's confidence in predicting the End Of Sentence (EOS) diminishes when under-translation occurs, serving as a mild penalty for under-translated candidates. Building upon this analysis, we propose employing the confidence of predicting EOS as a detector for under-translation, and strengthening the confidence-based penalty to penalize candidates with a high risk of under-translation.
Experiments on both synthetic and real-world data show that our method can accurately detect and rectify under-translated outputs, with minor impact on other correct translations.

\end{abstract}

\section{Introduction}
Neural Machine Translation (NMT) has made remarkable progress over the past years \citep{bahdanau2014neural,sutskever2014sequence,cho2014learning,vaswani2017attention}. Under-translation and over-translation are two typical problems in NMT, where under-translation means some words are mistakenly untranslated, and over-translation means some words are unnecessarily translated for multiple times \citep{tu-etal-2016-modeling,mi-etal-2016-coverage}.

Despite the rapid progress of NMT, the mechanisms behind the occurrence of under-translation and over-translation remain unclear, and these problems continue to be challenging obstacles faced by NMT systems \citep{he2020structure}. Prior efforts have primarily focused on modeling the explicit coverage of source content \citep{tu-etal-2016-modeling,mi-etal-2016-coverage,tu2017neural,zheng-etal-2018-modeling} or enforcing the attention-based source coverage during decoding \citep{wu2016googles}, without advancing the understanding of why NMT systems tend to overlook or repeat certain source words. \citet{zhao-etal-2018-exploiting,zhao2019addressing} empirically discovered that source words with a large translation entropy or those requiring reordering during translation are more likely to be ignored. However, this observation lacks a comprehensive explanation.

In this work, we conduct an in-depth analysis on the underlying cause of under-translation in NMT, providing an explanation from the perspective of decoding objective. Specifically, to gain deeper insights into the characteristics of under-translation and over-translation, we first generate synthetic data to simulate sentence-level and document-level translation scenarios, which allows us to automatically identify translation errors through predefined rules. The findings of sentence-level translation analysis indicate that words with high translation entropy, i.e., source words with a wide range of possible translations, are at a higher risk of being under-translated. In contrast, low-entropy words, whose translations are more predictable, are more likely to be over-translated, which aligns with the findings of \citet{zhao2019addressing}. From the document-level translation experiments, we further observe the phenomenon of sentence-level under-translation, where the last sentence containing many high-entropy words is typically omitted. Additionally, we notice that when under-translation occurs, the probability of predicting the End Of Sentence (EOS) is generally lower, suggesting that the model is unwilling to stop generation when the translation is incomplete. 

Although NMT imposes a penalty on predicting EOS, under-translation still occurs frequently, which can be explained from the perspective of decoding objective. NMT typically employs beam search for decoding, with an objective of finding the most probable sentence, or maximizing the log-probability normalized by sentence length \citep{wu2016googles}. Consequently, NMT has a strong incentive to ignore high-entropy words, as they have low translation probabilities that contradict the decoding objective. Moreover, we theoretically reveal that the lower EOS probability serves as a penalty for under-translated candidates, but the penalty often underweighs the benefits of dropping multiple high-entropy words, leading to the occurrence of under-translation.

Building upon this analysis, we propose enhancing the EOS penalty on under-translated candidates to prevent under-translation. Since the model's confidence in predicting EOS diminishes when under-translation occurs, we employ the prediction probability of EOS as a detector for under-translation. For beam search candidates with high risk of under-translation, we take the EOS probability as penalty and scale it to be proportional to the translation length. The detection ensures minimal interference with correct translations, and the scaling balances the impact of penalty with the benefits of dropping high-entropy source words, thereby more effectively preventing under-translations.

In summary, our contributions are:\vspace{3pt}\\
\indent $\vcenter{\hbox{\small{$\bullet$}}}$ We analyze the characteristics of under-translation based on our synthetic data, suggesting that under-translation is more likely to occur on challenging words or sentences.\vspace{3pt}\\
\indent $\vcenter{\hbox{\small{$\bullet$}}}$ We explain under-translation from the perspective of the decoding objective and theoretically reveal that the EOS probability serves as a penalty for under-translated candidates.\vspace{3pt}\\
\indent $\vcenter{\hbox{\small{$\bullet$}}}$ We propose enhancing the EOS penalty on beam search candidates at risk of under-translation. Experiments show that our method can accurately detect and rectify under-translated outputs, with minor impact on other correct translations. 

\section{Experiments on Synthetic Data}
\label{sec:2}
Under-translation and over-translation are two challenging problems in neural machine translation, with one primary difficulty being the lack of automatic metrics for them. Previous work has mainly relied on human annotators to identify translation errors, which is time-consuming. In this section, we introduce a method for constructing synthetic data to simulate sentence-level and document-level translation scenarios, which allows us to automatically identify under-translation and over-translation errors through predefined rules.

Specifically, we create synthetic data containing only three words: A, B, and C, representing low-entropy, mid-entropy, and high-entropy words, respectively. We establish a fixed translation probability from source words to target words, as shown in Table \ref{table:prob}. Notably, the most likely translation for each word is the word itself, but the probability distribution varies, with A being the sharpest and C being the smoothest. The optimal translation result should be a direct copy of the source, and any deviation indicates a translation error. Typically, a shorter translation compared to the source implies under-translation, while a longer translation implies over-translation.

\begin{table}[t]
\begin{center}
\begin{small}
\begin{tabular}{c|ccc}
\toprule
\diagbox{\textbf{src}}{\textbf{tgt}}   & \bf A & \bf B & \bf C \\
\midrule
\bf A & 80\% & 10\% & 10\% \\
\midrule
\bf B & 20\% & 60\% & 20\% \\
\midrule
\bf C & 30\% & 30\% & 40\% \\
\bottomrule 
\end{tabular}
\end{small}
\end{center}
\caption{The predefined translation probability from source words to target words.}
\label{table:prob}
\end{table}

\subsection{Sentence-level Translation}

\textbf{Dataset Construction.} To simulate sentence-level translation scenarios, we construct the training set with both source and target sides being multiple sentences, composed exclusively of the three words: A, B, and C. The length of each source sentence is randomly selected from the set $\{1,2,...,20\}$, with each source word being equally likely to be sampled from the set $\{A, B, C\}$. We translate the source words sequentially according to the probability distribution presented in Table \ref{table:prob}. Additionally, we simulate noise in the training set by introducing a 15\% chance of distortion for each word's translation, with an equal likelihood of either dropping the word or translating it twice. For the test set, the source side is constructed using the same methodology, with the target side being an exact replication of the source.

\vspace{5pt}
\noindent{}\textbf{Settings.} Following the above methodology, we construct 1 million sentence pairs for the training and 1,000 sentence pairs for the test. The model architecture is a scaled-down version of Transformer ($h_{dim} = 128, h_{ffn} = 256, heads = 2, layers = 2$). The number of training steps is 50,000. We apply beam search for decoding, with a beam size of 5 and a length penalty of 1. Other settings follow the standard configuration of Transformer-base \citep{vaswani2017attention}.

\vspace{5pt}
\noindent{}\textbf{Results.} We can automatically detect under-translation and over-translation errors by comparing the length of the source input and the model's decoding output. In the test set of 1,000 sentences, we identify a total of 53 instances of under-translation and 12 instances of over-translation, indicating that these issues, particularly under-translation, are significant and warrant attention. In Table \ref{table:wordlevel}, we present the word distribution in sentences with under-translation and over-translation errors. It can be observed that sentences with a higher proportion of the high-entropy word C are more prone to under-translation, which aligns with the findings of \citet{zhao2019addressing}. Additionally, we also discover that sentences with a higher proportion of the low-entropy word A are more likely to be over-translated.

\begin{table}[t]
\begin{center}
\setlength{\tabcolsep}{4pt}
\begin{small}
\begin{tabular}{c|ccc|ccc}
\toprule
&\multicolumn{3}{c|}{\bf Source} &\multicolumn{3}{c}{\bf Target} \\

 & \bf A & \bf B & \bf C  & \bf A & \bf B & \bf C\\
\midrule
\bf All & 33.7\% & 33.4\% & 32.9\% & 37.7\% & 34.8\% & 27.5\%  \\
\midrule
\bf Under & 28.5\% & 34.3\% & 37.2\% & 34.1\% & 37.8\% & 28.1\% \\
\midrule
\bf Over & 38.3\% & 27.4\% & 34.3\% & 45.5\% & 28.2\% & 26.3\% \\
\bottomrule 
\end{tabular}
\end{small}
\end{center}
\caption{Word distribution in under-translated and over-translated sentences.}
\label{table:wordlevel}
\end{table}

\subsection{Document-level Translation}
\textbf{Dataset Construction.} To simulate document-level translation scenarios, we further construct the dataset with source and target sides being a paragraph containing one or multiple sentences. We expand the vocabulary to $\{A,B,C,.\}$, where the new word `.' denotes the end of a sentence. The number of source sentences is randomly selected from the set $\{1,2,...,5\}$, and the length of source sentence is sampled from $\{1,2,...,20\}$. Besides the word-level noise, we also introduce a 15\% chance of distortion for each source sentence's translation, with an equal likelihood of either dropping the sentence or translating it twice.

\vspace{5pt}
\noindent{}\textbf{Settings.} We employ the same settings as the word-level scenario.

\vspace{5pt}
\noindent{}\textbf{Results.} In the document-level experiment, we examined sentence-level over-translation and under-translation errors by automatically detecting discrepancies in the number of periods (`.') between the source sentence and the model's decoding output. Analyzing a test set of 1,000 sentences, we find 27 instances of under-translation and 2 instances of over-translation at the sentence-level. As shown in Table \ref{table:senlevel}, under-translation errors can be categorized into three types: last sentence under-translation, penultimate sentence under-translation, and merging of the last two sentences, with the last sentence under-translation being the most common error scenario. Another observation is that the omitted sentences often contain a high proportion of high-entropy word C, making their translation more challenging. To systematically verify this, we further analyze the word distribution in the omitted sentences and find that the proportions of words A, B, and C are 30.2\%, 26.1\%, and 43.7\%, respectively. This finding aligns with the conclusion at the word level, suggesting that the last sentences containing many high-entropy words are more likely to be omitted.

\begin{table}[t]
\begin{center}
\setlength{\tabcolsep}{4pt}
\begin{small}
\begin{tabular}{cccc}
\toprule
\bf Type & \bf Source & \bf Output & \bf Count \\
\midrule
 Last & CABBAB. BCBCCC. & CABBAB. & 17   \\
\midrule
 Penultimate & CCBCA. CBAAA. & CBAAA. & 4  \\
\midrule
 Merge & ABCCCA. BCBBAA. & ABBBAA. & 6  \\
\bottomrule 
\end{tabular}
\end{small}
\end{center}
\caption{Examples of sentence-level under-translation errors. For brevity, we provide only the last two source sentences and their translation outputs. `Count' denotes the number of this error type in the test set.}
\label{table:senlevel}
\end{table}

\section{Approach}

As observed in the previous section, under-translation poses a significant challenge in NMT, particularly when high-entropy words and difficult sentences are involved. To provide a comprehensive understanding of this phenomenon, we offers an intuitive explanation and theoretical analysis from the perspective of the decoding objective in Section \ref{sec:31}. Following this, we investigate the relationship between the EOS probability and under-translation in Section \ref{sec:32}, and then design the EOS penalty to prevent under-translation in Section \ref{sec:33}.

\subsection{Effect of Decoding Objective}
\label{sec:31}

When humans engage in translation, their objective is to accurately convey the meaning of the source text using words in the target language. In contrast, when machines perform translation, their objective is expressed through a mathematical formula:
\begin{equation}
\label{eq:obj}
\max_{Y}\ \frac{\log P_{\theta}(Y|X)}{|Y|^{\alpha}},
\end{equation}
where $X,Y$ represent the source and target sentence, respectively, and $\alpha$ denotes the length penalty. We argue that this discrepancy results in deviations between machine translation outputs and human expectations, with a typical outcome being the under-translation of high-entropy words and difficult sentences.

Maximizing the length-normalized log-probability implies that the output words should have the highest log-probabilities in the average sense. This inherently encourages the model to disregard words that fall below average, which contradicts the intention of producing a comprehensive translation. The deterrent to under-translation is the potential incoherence of the translation if certain words are omitted, which reduces the overall translation probability. However, there are instances where the incomplete translation of some words does not significantly impact other parts of the translation. In such cases, the exclusion of high-entropy words and difficult sentences can be particularly attractive to the model.

The dominant type of sentence-level under-translation, the omission of the last sentence, serves as an example. Suppose the model generates two candidates when translating a document: $Y_{pre}$, which excludes the last sentence, and $Y_{pre:last}$, a comprehensive translation that includes the last sentence $Y_{last}$. The model's choice between these candidates depends on the comparison of their decoding objectives, i.e., whether the following condition is satisfied:
\begin{equation}
\label{eq:2}
\frac{\log P_{\theta}(Y_{pre:last}|X)}{|Y_{pre:last}|^{\alpha}} > \frac{\log P_{\theta}(Y_{pre}|X)}{|Y_{pre}|^{\alpha}}.
\end{equation}
Given that $Y_{last}$ represents a single sentence and $Y_{pre}$ refers to all preceding sentences within a document, it is reasonable to assume that their length ratio $\lambda=\frac{|Y_{pre}|}{|Y_{last}|}$ is close to 0. This assumption enables us to approximate $(1+\lambda)^{\alpha} \approx 1+\alpha\lambda$. Leveraging this approximation, we can transform the above condition into the following inequality (see Appendix \ref{app:proof} for a detailed derivation):
\begin{equation}
\begin{aligned}
\label{eq:3}
& - \frac{\log P_{\theta}(eos|X,Y_{pre\backslash eos})}{|Y_{last}|} > \\
&  \alpha \cdot  \frac{\log P_{\theta}(Y_{pre}|X)}{|Y_{pre}|}- \frac{\log P_{\theta}(Y_{last}|X,Y_{pre})}{|Y_{last}|}.
\end{aligned}
\end{equation}

It can be observed that the model's decision to exclude the last sentence is primarily based on a comparison between its average log-probability and the product of the average log-probability of previous sentences and a length penalty. The log-probability of the EOS token also acts as a penalty against under-translation. Intuitively, the model's confidence in predicting EOS should diminish when the translation is incomplete, thereby enhancing the penalty for under-translation. Although the above analysis is specifically for the omission of the last sentence, it is generally applicable to other cases of sentence-level under-translation, with an additional factor involved, namely the impact of missing a middle sentence on the conditional probabilities of subsequent sentences.

\subsection{EOS Probability Analysis}
\label{sec:32}

\begin{figure}[t]
\centering
\begin{minipage}{.25\textwidth}
  \centering
  \includegraphics[width=1.\linewidth]{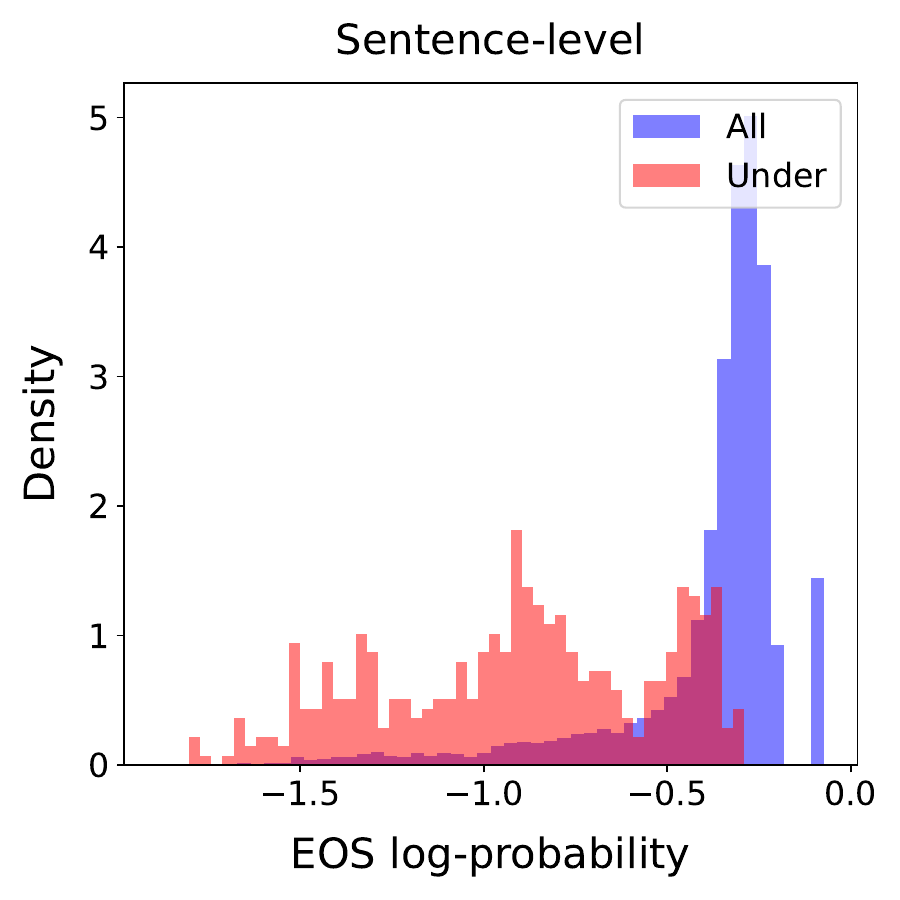}
\end{minipage}%
\begin{minipage}{.25\textwidth}
  \centering
  \includegraphics[width=1.\linewidth]{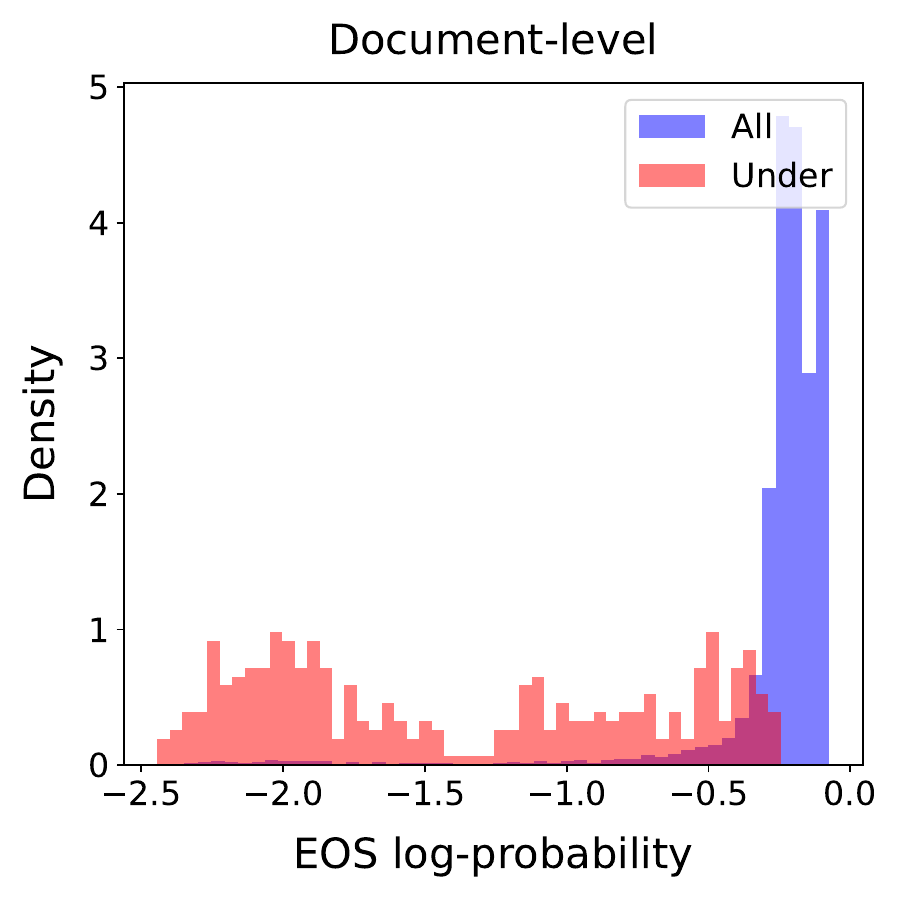}
\end{minipage}
  \caption{Comparison of EOS log-probability distribution: under-translated sentences vs. all sentences on sentence-level (left) and document-level (right) synthetic test sets.}
  \label{fig:eosdis}
\end{figure}

\begin{figure}[t]
  \begin{center}
    \includegraphics[width=.9\columnwidth]{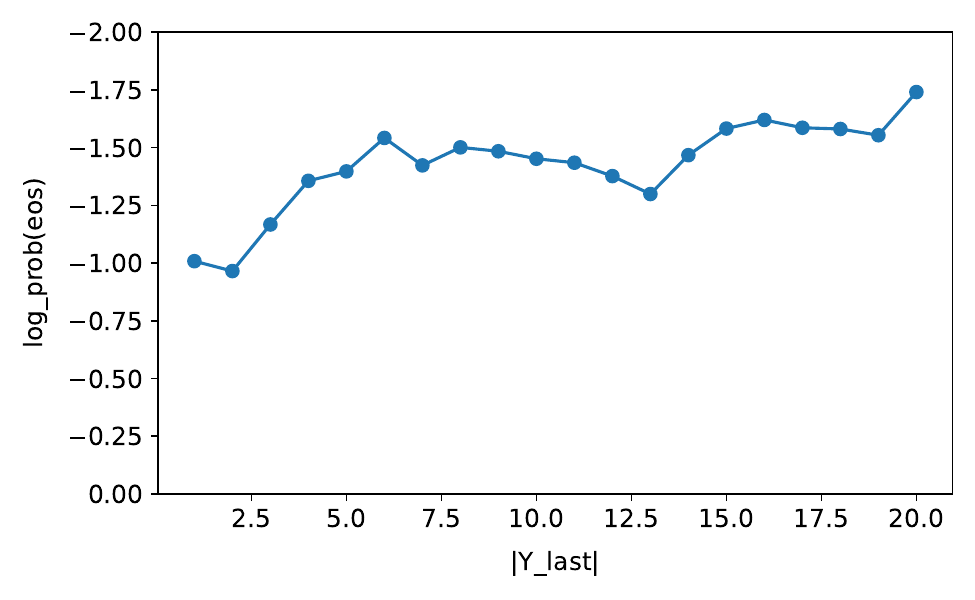}
    \caption{Average EOS log-probability $\log P_{\theta}(eos|X,Y_{pre\backslash eos})$ in relation to number of under-translated words $|Y_{last}|$ on document-level synthetic test set.}
    \label{fig:eoscurve}
  \end{center}
\end{figure}
The above analysis suggests that the log-probability of EOS acts as a penalty against under-translation. Even when a sentence contains many high-entropy words that make it challenging to translate, the model will still favor the complete translation as long as it assigns a low EOS log-probability to the imcomplete translation.

Therefore, we first investigate whether NMT tends to assign lower end-of-sentence probabilities to unfished translations. We conduct experiments on both sentence-level and document-level synthetic test sets, comparing the distribution of EOS log-probabilities between under-translated sentences and all sentences within the test set. The results are presented in Figure \ref{fig:eosdis}. As can be seen, on both datasets, the EOS log-probabilities of under-translated sentences are significantly lower, suggesting that the model is unwilling to stop generation when the translation is incomplete. This effect is particularly evident when sentence-level under-translation takes place within the document-level dataset, indicating that the decrease in EOS log-probability is also influenced by the number of under-translated words.

Equation \ref{eq:3} implies that the log-probability of EOS needs to scale linearly with the number of under-translated words in order to effectively prevent under-translation. Therefore, we delve deeper into the relationship between EOS log-probability and the number of under-translated words to verify whether this requirement is fulfilled. On the document-level synthetic test set, we compute the average EOS log-probability for each length of under-translated sentences. The results are shown in Figure \ref{fig:eoscurve}. As observed, although the EOS log-probability and the number of under-translated words generally display a positive correlation, the rate of change does not reach linearity. Therefore, the EOS penalty tends to underweigh the benefits of dropping multiple high-entropy words, leading to the occurrence of under-translation.

\subsection{Enhancing the EOS Penalty}
\label{sec:33}

Based on the above analysis of EOS probability, we have two key findings to help address the under-translation problem. First, the model's confidence in predicting EOS diminishes when under-translation occurs, enabling us to utilize the model's predicted probability of EOS to assess the risk of under-translation. Second, the log-probability of EOS acts as a penalty against under-translation, but its effect is outweighed by the benefits of dropping multiple high-entropy words. Therefore, we can further enhance the EOS penalty for candidates with a high risk of under-translation, leading to a more precise and effective prevention of under-translations.

To utilize the EOS probability for under-translation detection, the most straightforward approach is to establish a threshold and label sentences with an EOS log-probability below this threshold as prone to under-translation. However, as illustrated in Figure \ref{fig:eosdis}, the distribution of EOS log-probability varies across different datasets, so a fixed threshold may not perform well universally. We empirically find that a comparative metric offers more consistent performance across different datasets. Specifically, we compare the EOS token against its closest competitor, $w_2$, which is the second-highest ranked word in the final step. A translation is considered prone to under-translation if the EOS probability does not significantly exceed that of $w_2$, failing to satisfy the following condition:
\begin{equation}
\label{eq:cond}
\log P_{\theta}(eos|X,Y) - \log P_{\theta}(w_2|X,Y) > \tau,
\end{equation}
where $\tau=1$ is a hyperparameter that exhibits consistent performance across different datasets.

Upon detecting beam search candidates at risk of under-translation, we can enhance their EOS penalty to prevent potential under-translation errors. A straightforward approach involves amplifying the log-probability of EOS by a constant factor. However, for short sentences, this method may lead to the decoding objective being dominated by the EOS probability, which can negatively impact the overall translation quality. 

To address this, we modify the approach by scaling the EOS penalty based on the translation length, which avoids excessive effect on short sentences while ensuring a sufficient penalty for longer sentences. Another consideration is that when the decoded sentence is relatively long, the beam search process generally retains candidates with consistent prefixes due to the limited beam size. In such cases, it is unnecessary to fully amplify the EOS penalty based on the entire translation length. Instead, we can truncate the length by setting an upper limit. In summary, the final score for beam search candidates are adjusted as follows:
\begin{equation*}
\begin{cases}
      \frac{\log P_{\theta}(Y|X)}{|Y|^{\alpha}} , & \text{if meets Cond.}\ref{eq:cond} \\
      \frac{\log P_{\theta}(Y|X) + \lambda\cdot \log P_{\theta}(eos|X,Y)}{|Y|^{\alpha}}, & \text{otherwise}
      \end{cases},
\end{equation*}
\begin{equation}
\label{eq:5}
 \lambda = \beta \cdot \min(|Y|,L),
\end{equation}
where $\beta=0.4$ and $L=20$ are hyperparameters that control the penalty weight $\lambda$.

\section{Experiments}
\subsection{Settings} 
\textbf{Datasets.} 
We validate our method on both sentence-level and document-level benchmarks using synthetic and real-world translation data. The construction details of the synthetic data have been described in section \ref{sec:2}. For real-world translation, we use the test set of WMT22 Chinese-to-English translation (WMT22 Zh-En, 1,875 sentence pairs) as the sentence-level translation benchmark \citep{kocmi-etal-2022-findings}. For document-level translation, we combine sentences from the WMT22 Zh-En test set into documents, stopping when the length exceeds 200. This results in a document-level translation benchmark comprising 328 document pairs.

\vspace{5pt}
\noindent{}\textbf{Models.} For synthetic data, the model architecture is a scaled-down version of Transformer ($h_{dim} = 128, h_{ffn} = 256, heads = 2, layers = 2$). The training details have been described in section \ref{sec:2}. For real-world translation, we employ large language models (LLMs) as they demonstrate good performance in both sentence-level and document-level translation tasks \citep{jiao2023chatgpt,wang2023document,wu2024adapting}. Our foundation model is LLaMA2-7B \citep{touvron2023llama}, which we fine-tune for 1 epoch using the Alpaca instruction dataset \citep{selfinstruct,alpaca} and Chinese-English test sets from WMT17-20. We also augment the translation data by concatenating three consecutive sentences as a document, thereby equipping the final model with both sentence-level and document-level translation capabilities. Other training hyperparameters are the same as Alpaca-7B \citep{alpaca}.

\vspace{5pt}
\noindent{}\textbf{Decoding.} The prompt for decoding is ``Translate the following sentences from Chinese to English''. Unless otherwise specified, the beam size is 5 and the length penalty is 1.0.

\subsection{Results on Synthetic Data}

First, we conduct experiments on synthetic data to investigate the impact of enhancing the EOS penalty on under-translation and over-translation. In sentence-level NMT, we measure the proportion of sentences with word-level under-translation and over-translation errors. In document-level NMT, we do not consider word-level errors and only measure the proportion of translations with sentence-level under-translation and over-translation errors. The experimental results are shown in Table \ref{table:syn}.

\begin{table}[t]
\begin{center}
\begin{small}
\begin{tabular}{lcc}
\toprule
\bf System & \bf Under & \bf Over \\
\midrule
 Sentence-level NMT & 5.3\% & 1.2\% \\
\ \ + EOS penalty & 2.3\% & 1.9\% \\
\midrule
 Document-level NMT & 2.7\% & 0.2\% \\
\ \ + EOS penalty & 1.5\% & 0.3\% \\
\bottomrule 
\end{tabular}
\end{small}
\end{center}
\caption{The proportions of under-translation and over-translation on synthetic test sets.}
\label{table:syn}
\end{table}

As can be seen, the main issue faced by the model is under-translation, with far fewer cases of over-translation in comparison. After enhancing the EOS penalty, the under-translation problem significantly decreases in both sentence-level and document-level NMT. The side effect is an increase in over-translation, but this is relatively minor compared to the reduction in under-translation. In document-level NMT, we have categorized the under-translation errors in synthetic data into three tpyes: last, penultimate, and merge, as shown in Table \ref{table:senlevel}. Our method primarily resolves last sentence under-translation, with 10 out of the 12 reduced under-translation errors belonging to this type.

\subsection{Results on Real-world Data}
For real-world translation, we conduct human analysis to identify under-translation and over-translation errors. We take the first 300 sentences from the sentence-level test set and the first 100 documents from the document-level test set, and measure the proportion of under-translation and over-translation on these subsets. The results are shown in Table \ref{table:zhen}.

\begin{table}[t]
\begin{center}
\begin{small}
\begin{tabular}{lcc}
\toprule
\bf System & \bf Under & \bf Over \\
\midrule
 Sentence-level NMT & 9.7\% & 2.3\% \\
\ \ + EOS penalty & 6.7\% & 2.3\% \\
\midrule
 Document-level NMT & 14.0\% & 4.0\% \\
\ \ + EOS penalty & 10.0\% & 5.0\% \\
\bottomrule 
\end{tabular}
\end{small}
\end{center}
\caption{The proportion of under-translation and over-translation on subsets of WMT22 Zh-En test sets.}
\label{table:zhen}
\end{table}

We can observe that under-translation is also a significant problem in real-world translation scenarios. Over-translation, also referred to as hallucination, is less common in comparision. Enhancing the EOS penalty effectively mitigates the occurrence of under-translation in both sentence-level and document-level NMT. In contrast, we have only observed one instance of over-translation resulting from the enhanced EOS penalty.

Upon further examination of the variations in translations, we find that our method is particularly prominent in terms of precision, affecting only about 10\% of the translations. After enhancing the EOS penalty, the modifications in translations mainly fall into three types: adding punctuation at the end, paraphrasing of translated sentences, and retrieving missing translations, as illustrated in Table \ref{table:diff}. In the 300 sentences we extracted, the number of modifications in these three types are 8, 6, and 9, respectively. In the 100 documents we extracted, they are 3, 1, and 6, respectively. As can be seen, the modifications primarily focus on mitigating under-translations, with minor impact on correct translations. We present the details of these translation differences in Appendix \ref{app:diff}. 

\begin{table}[t]
\begin{center}
\begin{small}
\begin{tabular}{ccc}
\toprule
\bf Type & \bf w/o EOS penalty & \bf w/ EOS penalty \\
\midrule
1 & I didn't get it & I didn't get it.\\
\midrule
\multirow{2}{*}{2} & 13 yuan for a&13 yuan for one\\& shrimp dumpling? & shrimp dumpling?\\
\midrule
\multirow{3}{*}{3} &Or how long does &Or how long does the\\&the restaurant need & restaurant need to prepare?\\&to prepare? &Can you help me ask?\\
\bottomrule 
\end{tabular}
\end{small}
\end{center}
\caption{Illustration of three types of modifications: 1. adding punctuation at the end, 2. paraphrasing of translated sentences, and 3. retrieving missing translations.}
\label{table:diff}
\end{table}

\subsection{Effect of Detection}

Our method particularly prominent in terms of precision on mitigating under-translations, which can be attributed to the incorporation of an under-translation detection module (Equation \ref{eq:cond}). The detection allows us to apply the EOS penalty selectively to sentences that exhibit a high risk of under-translation, thus having a minor impact on other correct translations. 

The effect of the under-translation detection module is illustrated in Table \ref{table:detect}. It can be observed that incorporating the EOS detection module results in only a slight reduction in the number of retrieved under-translations. The advantage lies in that it significantly reduces the impact on other originally correct translations, thereby enhancing the precision of under-translation rectification.

\begin{table}[t]
\begin{center}
\begin{small}
\begin{tabular}{ccccc}
\toprule
\multirow{2}{*}{\bf Type} & \multicolumn{2}{c}{\bf Sentence-level} & \multicolumn{2}{c}{\bf Document-level} \\
 &\bf w/ det.&\bf w/o det.&\bf w/ det.&\bf w/o det.\\
\midrule
1& 8 & 14 & 3 &4\\
\midrule
2& 6 & 13 & 1 &3\\
\midrule
3& 9 & 11 & 6 &6\\
\bottomrule 
\end{tabular}
\end{small}
\end{center}
\caption{The number of three types of modifications measured on subsets of WMT22 Zh-En test sets. `det.' is the abbreviation of detection.}
\label{table:detect}
\end{table}

\subsection{Types of Under-translation}

In real-world scenarios, we found that the categories of under-translation errors are not entirely consistent with those summarized in the synthetic data, but there are similarities. In both sentence-level and document-level NMT, we divide under-translation errors into three types according to their occurrence position: begin, middle, and end. In this context, `begin' refers to the first few (1-3) words of a sentence or the document's first sentence, and `end' exhibits a similar meaning. We present the distribution of under-translation errors in Table \ref{table:undertype}.

\begin{table}[t]
\begin{center}
\begin{small}
\begin{tabular}{lccc}
\toprule
\bf System & \bf Begin & \bf Middle & \bf End \\
\midrule
 Sentence-level NMT  & 3 & 12 & 14\\
\ \ + EOS penalty & 3 & 9& 8\\
\midrule
 Document-level NMT & 1 & 6 & 7 \\
\ \ + EOS penalty &  1& 6& 3\\
\bottomrule 
\end{tabular}
\end{small}
\end{center}
\caption{The number of under-translation errors in different occurrence positions measured on subsets of WMT22 Zh-En test sets.}
\label{table:undertype}
\end{table}

We can see that under-translation is most likely to occur at the end, where the last sentence accounts for only a small portion of the document but contributes to half of the under-translation errors. This observation is consistent with our earlier findings in synthetic data and aligns with our theoretical analysis. Under-translation at the end is also the primary focus of our method. In sentence-level translation, it eliminates 9 under-translated sentences, with 3 in the middle position and 6 at the end. In document-level translation, our method detects 6 out of 7 under-translations at the end and corrects 4, while the remaining two still exhibit some under-translation after modification.

A limitation of our method is its weaker ability to handle under-translation at the begin and middle stages. Our method modifies final scores and rankings of beam search candidates but cannot influence the early decoding process. Due to the limited beam size, candidates containing complete translations are already dropped during beam search, so we are unable to resolve these under-translation errors by modifying scores of final candidates.

\subsection{Impact on Translation Quality}

\begin{table}[t]
\begin{center}
\begin{small}
\begin{tabular}{lcc}
\toprule
\bf System & \bf BLEU  \\
\midrule
 Sentence-level NMT  & 23.34\\
\ \ + EOS penalty & 23.47 \\
\midrule
 Document-level NMT & 24.30   \\
\ \ + EOS penalty &  24.32\\
\bottomrule 
\end{tabular}
\end{small}
\end{center}
\caption{BLEU scores on WMT22 Zh-En test sets.}
\label{table:metric}
\vspace{-0.5em}
\end{table}

In the previous sections, our evaluation has been focused on the effects of our method on under-translation and over-translation, but its overall impact on translation quality remains unclear. In this section, we further examine the impact of our method on the BLEU \citep{papineni-etal-2002-bleu}, the most widely used evaluation metric in machine translation. We measure the sacreBLEU \citep{post-2018-call} and report the results in Table \ref{table:metric}.

The results indicate that addressing the under-translation problem can lead to a positive improvement in the evaluation metric. However, this improvement is relatively moderate, which be explained by two main factors. First, the errors resulting from under-translation do not comprise a substantial portion of the entire test set, thus limiting the overall improvement from resolving under-translation. Second, as we have analyzed, sentences with increased complexity and a higher number of high-entropy words are more prone to under-translation. Therefore, the retrieved translations may not achieve the same level of accuracy as other translations in the test set, resulting in a limited improvement in the metrics.

\subsection{Comparison with Other Penalties}

\begin{table}[t]
\begin{center}
\begin{small}
\begin{tabular}{clccc}
\toprule
\multicolumn{2}{c}{\bf Penalty} &\bf $\Delta_{\text{Under}}/\Delta_{\text{All}}$ & \bf $\Delta_{\text{BLEU}}$ & \bf $\Delta_{\text{Tokens}}$ \\
\midrule
\multirow{3}{*}{Length}&$\alpha$=1.5 &  2$/$14 &+ 0.01&236 \\
&$\alpha$=2.0 &  4$/$26  &+ 0.09&445\\
&$\alpha$=2.5 &  5$/$48  &- 0.18&1321\\
\midrule
\multirow{3}{*}{Coverage}&$\beta$=0.05&5$/$39&+ 0.11&521\\
&$\beta$=0.1 &  9$/$77 &+ 0.10 &1112\\
&$\beta$=0.2 &   11$/$139 &- 0.21&2092\\
\midrule
\multicolumn{2}{c}{ EOS Penalty}&9$/$24&+ 0.13&565  \\
\bottomrule 
\end{tabular}
\end{small}
\end{center}
\caption{Effect of different penalties on the WMT22 Zh-En test set. $\Delta_{\text{BLEU}}$ means the BLEU difference compared to the NMT baseline. $\Delta_{\text{Tokens}}$ represents the difference in the number of tokens. $\Delta_{\text{Under}}$ and $\Delta_{\text{All}}$ are the number of resolved under-translations and all different sentences on the subset of 300 sentences.}
\label{table:penalty}
\end{table}

In addition to the proposed EOS penalty, \citet{wu2016googles} has introduced the length penalty and coverage penalty to penalize under-translation, where we have incorporated a length penalty of 1.0 in our method. In this section, we further explore the effect of length penalty and coverage penalty on under-translation, and compare them with the EOS penalty, as illustrated in Table \ref{table:penalty}. 

We find that both the length penalty and coverage penalty exhibit some ability to resolve under-translation, with the effect of the coverage penalty being more significant. However, their precision on addressing under-translation ($\Delta_{\text{Under}}/\Delta_{\text{All}}$) is relatively lower, affecting more unrelated translations. Specifically, we note that both penalties tend to convert translations into synonymous but longer forms, sometimes even causing hallucinations, which is reflected by the increased number of tokens $\Delta_{\text{Tokens}}$.

\section{Related Work}
Research on the under-translation problem in NMT can be roughly divided into three categories: modeling source coverage, exploring translation entropy, and integrating penalties during the decoding.

\vspace{5pt}
\noindent{}\textbf{Source Coverage.} \citet{tu-etal-2016-modeling} pointed out the problems of under-translation and over-translation in neural machine translation and proposed a method for tracking source coverage vector to adjust the decoding of model. Similarly, \citet{mi-etal-2016-coverage,tu2017neural,zheng-etal-2018-modeling} focused on modeling the coverage of source content. \citet{mi-etal-2016-coverage} introduced a coverage embedding vector to track the coverage of source words. \citet{tu2017neural} proposed to reconstruct the source sentence, thereby ensuring that all source information is included in the target side. \citet{zheng-etal-2018-modeling} addressed under-translation by making the model aware of translated and untranslated contents. 

\vspace{5pt}
\noindent{}\textbf{Translation Entropy.} \citet{zhao-etal-2018-exploiting,zhao2019addressing} discovered that source words with a large translation entropy or those requiring reordering during translation are more likely to be ignored, and proposed pre-ordering the source sentence or converting high-entropy words into special tokens to prevent under-translation. \citet{liu-etal-2022-part,chen-etal-2023-multifaceted} further employed entropy to assess translation difficulty, using this information to create test sets or enhance evaluation metrics.

\vspace{5pt}
\noindent{}\textbf{Decoding Penalty.} \citet{wu2016googles} proposed the length penalty and coverage penalty to penalize under-translation. The length penalty adjusts the log-probability score based on sentence length, reducing bias towards short sentences. The coverage penalty encourages source-target attention to cover all source words to prevent under-translation.

\section{Conclusion}
In this paper, we conduct an in-depth study of the under-translation problem in NMT, explaining and addressing this problem from the perspective of the decoding objective. Our analysis reveals that under-translation is more likely to occur on challenging words or sentences. We explain it from the perspective of the decoding objective and propose enhancing the EOS penalty on under-translated candidates to prevent under-translation. Experimental results show that our method effectively mitigates the occurrence of under-translation, with only a minor impact on other correct translations.

\section{Limitations}
This paper investigates the underlying causes of under-translation in NMT and proposes a solution based on the End-of-Sentence (EOS) penalty. This is a preliminary attempt from the perspective of the decoding objective, and there are limitations that warrant future improvements.

First, our method modifies the final scores and rankings of beam search candidates but cannot influence the early decoding process. Therefore, when correct and complete translations are already dropped during the early search process, our method cannot retrieve the missing translation. Overcoming this limitation may be possible by identifying additional signals strongly associated with under-translation during decoding and designing corresponding penalties.

Second, the period (`.') plays a similar role with EOS that indicates the end of a sentence. In sentences ending with a period, the EOS probability is typically high, making it difficult to identify under-translation. A possible future direction is to construct a vocabulary of punctuation marks, taking into account words like periods that indicate the end of a sentence when designing the penalty.

Lastly, our synthetic data construction does not fully simulate real translation data. For instance, we only translate source words sequentially when constructing the target-side translation, neglecting potential reordering and dependencies between target words. A more sophisticated construction of synthetic data may help us further reveal the characteristics of under-translation and design better decoding objectives.

\bibliography{custom}
\appendix
\onecolumn
\section{Proof}
\label{app:proof}

In this section, we present the derivation process from Equation \ref{eq:2} to Equation \ref{eq:3}. To begin, we redefine the notation to avoid ambiguity. In the following, we will use $Y_{pre}$ and $Y_{last}$ to represent sentences in a linguistic context without the EOS token, and use $\overline{Y_{pre}}$ and $\overline{Y_{last}}$ to denote sentences that include the EOS token. In this way, Equation \ref{eq:2} can be rewritten as:
\begin{equation*}
\frac{\log P_{\theta}(Y_{pre}|X) + \log P_{\theta}(Y_{last}|X,Y_{pre})+\log P_{\theta}(eos|X,Y_{pre:last})}{(|Y_{pre}|+|Y_{last}|)^{\alpha}} > \frac{\log P_{\theta}(Y_{pre}|X) + \log P_{\theta}(eos|X,Y_{pre})}{|Y_{pre}|^{\alpha}}.
\end{equation*}
Let $\lambda = \frac{|Y_{last}|}{|Y_{pre}|}$ represent the length ratio of the last sentence to the preceding sentences. The above equation can be transformed to:
\begin{equation*}
\frac{\log P_{\theta}(Y_{pre}|X) + \log P_{\theta}(Y_{last}|X,Y_{pre})+\log P_{\theta}(eos|X,Y_{pre:last})}{(1+\lambda)^{\alpha}} > {\log P_{\theta}(Y_{pre}|X) + \log P_{\theta}(eos|X,Y_{pre})}.
\end{equation*}
Since $Y_{last}$ represents a single sentence and $Y_{pre}$ refers to all preceding sentences within a document, it is reasonable to assume that their length ratio $\lambda$ is close to 0. Thus, we can approximate $(1+\lambda)^{\alpha} \approx 1+\alpha\lambda$, which is the only approximation made in the derivation. Next, we multiply both sides of the equation by $1+\alpha\lambda$ and cancel out common terms, yielding:
\begin{equation*}
\begin{aligned}
\log &P_{\theta}(Y_{last}|X,Y_{pre})+\log P_{\theta}(eos|X,Y_{pre:last}) > \\
&\alpha\lambda\cdot (\log P_{\theta}(Y_{pre}|X) + \log P_{\theta}(eos|X,Y_{pre})) + \log P_{\theta}(eos|X,Y_{pre}).\\
\end{aligned}
\end{equation*}
Further simplifying the equation with the notation $\overline{Y_{pre}}$ and $\overline{Y_{last}}$, we obtain:
\begin{equation*}
\log P_{\theta}(\overline{Y_{last}}|X,Y_{pre}) > \alpha\lambda\cdot \log P_{\theta}(\overline{Y_{pre}}|X) + \log P_{\theta}(eos|X,Y_{pre}).
\end{equation*}
Dividing each side by $|Y_{last}|=\lambda \cdot |Y_{pre}|$, we get:
\begin{equation*}
\frac{\log P_{\theta}(\overline{Y_{last}}|X,Y_{pre})}{|Y_{last}|} > \alpha\cdot \frac{\log P_{\theta}(\overline{Y_{pre}}|X)}{|Y_{pre}|} + \frac{\log P_{\theta}(eos|X,Y_{pre})}{|Y_{last}|}.
\end{equation*}
Finally, by reorganizing the terms and reverting to the initial notation, we can get Equation \ref{eq:3} as presented in the main text.
\newpage
\section{Translation Differences}
\label{app:diff}
\begin{table}[H]
\begin{center}
\resizebox{\linewidth}{!}{
\begin{small}
\begin{tabular}{ccccc}
\toprule
\bf Line & \bf Type & \bf Reference & \bf w/o EOS penalty & \bf w/ EOS penalty \\
\midrule
\multirow{2}{*}{2} & \multirow{2}{*}{1}& So that such a thing won’t& So as not to happen again & So as not to happen again.\\
&&happen again.&&\\
\midrule
\multirow{3}{*}{3} &\multirow{3}{*}{3} & And how much longer does it take& Or how long does the restaurant & Or how long does the restaurant\\
&&for the restaurant to prepare it?&need to prepare?&need to prepare?\\
&&Could could help me ask them?&&Can you help me ask?\\
\midrule
\multirow{2}{*}{23} & \multirow{2}{*}{2}& Washing hands with air:& Washing hands with air: & Washing hands with air:\\
&&save water in a cool way&Provincial water is cool&Water shortage and scorching heat\\
\midrule
\multirow{2}{*}{35} & \multirow{2}{*}{1}& If so, please place an& If so, please order as soon & If so, please order as soon\\
&&order soon.&as possible&as possible.\\
\midrule
\multirow{2}{*}{36}&\multirow{2}{*}{1}&We don’t have it at the moment,&Not now, thank you&Not now, thank you.\\
&&thank you. (not accurate)&&\\
\midrule
\multirow{2}{*}{37} &\multirow{2}{*}{3}&He asked me to tell you& He called me to tell you & He called me to tell you\\
&&his location \#PRS\_ORG\#.&where he is \#PRS\_ORG&where he is \#PRS\_ORG\#\\
\midrule
\multirow{2}{*}{63} &\multirow{2}{*}{2}&There is only one 13-yuan shrimp & 13 yuan for a shrimp dumpling? & 13 yuan for one shrimp dumpling?\\
&&dumpling!?&&\\
\midrule
\multirow{2}{*}{68} &\multirow{2}{*}{3}&Because I clearly saw the code & Because the code I input before & Because I saw that the code\\
&&was valid when I entered it.&was valid&I input before was valid\\
\midrule
\multirow{3}{*}{69} &\multirow{3}{*}{3}&I will say my conclusion first;& In conclusion, I think that & In conclusion, I think that "Four\\
&&I think Only Fools Rush In is&"Four Seas" is a comprehensive&Seas" is a comprehensive bad\\
&&utterly a terrible film in any sense.&bad film.&film. From various aspects, it is.\\
\midrule
\multirow{4}{*}{102} &\multirow{4}{*}{3}& CN1 connector CN1 signal I/O&CN1 connector CN1 signal I/O&CN1 connector CN1 signal I/O\\
&&connection cable CN1 port&Connector CN1 Connector CN1&Connector CN1 Connector CN1\\
&&terminal block, brake resistance&Connector&Connector Used End Terminal \\
&&&&CN1 Connector Brake Resistor\\
\midrule
\multirow{5}{*}{133} &\multirow{5}{*}{3}& Inspection shaft concrete module&Inspection of the production and&Inspection of the production and\\
&&production and processing &processing equipment of the&processing equipment of the\\
&&equipment, the assembly site for&inspection well concrete block&inspection shaft concrete module\\
&&Zhengzhou shaft wall brick&module:&Zhengzhou inspection shaft brick \\
&&module equipment&&module production line:\\
\midrule
148&2& ...... the vibration effect. & ...... the effectiveness of vibration. &...... the resonance effect. \\
\midrule
\multirow{2}{*}{180} &\multirow{2}{*}{3}& The product is good, worth&It is very good. It is worth&It is very good. It is worth\\
&& buying, really nice.& buying.& buying. It is very good.\\
\midrule
\multirow{3}{*}{184} &\multirow{3}{*}{2}& The big and clear screen is friendly&The screen is large, clear,&The screen is large and clear,\\
&&to my eyes, and no abnormality&the eyes are not tired, and there&and the eyes are not tiring. There\\
&&has been detected so far.&are no other abnormalities found&are no other abnormalities so far.\\
\midrule
\multirow{2}{*}{198} &\multirow{2}{*}{2}& Rice with roast pork is now&Cooking five-flower meat&Cooking five-flower meat\\
&&rice with pork and kimchi.&with five-flower meat&with five-flower vegetables\\
\midrule
\multirow{2}{*}{200} &\multirow{2}{*}{1}& It’s worrying, how long do I have&It is worrying that it will&It is worrying that it will\\
&& to wait before it arrives?&take so long to complete&take so long to complete.\\
\midrule
\multirow{2}{*}{204} &\multirow{2}{*}{1}& The restaurant said the food had been&The restaurant said it had cooked &The restaurant said it had cooked\\
&& cooked for more than half an hour.&the food for more than half an hour&the food for more than half an hour.\\
\midrule
\multirow{2}{*}{238} &\multirow{2}{*}{2}& The food is cold, and food can &The food has cooled down, &The food has cooled down.\\
&&go bad.& and the food may spoil&The food may spoil.\\
\midrule
241 &1& I didn’t receive it. &I didn't get it &I didn't get it.\\
\midrule
\multirow{3}{*}{245} &\multirow{3}{*}{3}&...... make abduction an action with&...... trafficking in human beings&...... trafficking in human beings will\\
&&no benefit, high risk and cost, then&  will not occur again and again.& not occur frequently because it is\\
&&such crimes will not happen again.&& unprofitable, high-risk and high-cost.\\
\midrule
\multirow{5}{*}{274} &\multirow{5}{*}{3}&Yang Haodong publicized the&Yang Haodong expounded the&Yang Haodong expounded the spirit\\
&&of the Sixth Plenary Session of the &spirit of the 19th CPC National&of the 19th National Congress\\
&&19th Central Committee of the CPC&Congress and the provincial&of the Communist Party of China\\
&&and the provincial congress of&party congress&and the spirit of the provincial\\
&&Party representatives.&&party congress at Ma Lingshan\\
\midrule
\multirow{2}{*}{288} &\multirow{2}{*}{1}&Because the delivery time is 30&Because the delivery time is&Because the delivery time is\\
&&minutes faster than what is displayed.&faster than 30 minutes&faster than 30 minutes.\\
\midrule
289&1&Nothing else, thanks.&No, thank you&No, thank you.\\
\bottomrule 
\end{tabular}
\end{small}
}
\end{center}
\caption{Translation differences after enhancing the EOS penalty (WMT22 Zh-En test set, Line 1-300).}
\label{table:diffsenall}
\end{table}

\begin{table}[H]
\begin{center}
\resizebox{\linewidth}{!}{
\begin{small}
\begin{tabular}{ccccc}
\toprule
\bf Line & \bf Type & \bf Reference & \bf w/o EOS penalty & \bf w/ EOS penalty \\
\midrule
\multirow{3}{*}{165-170} &\multirow{3}{*}{3}&...... I live in Shangshui, and why&...... I don't know why the order&...... I don't know why the order\\
&&is the order in Haihui Garden?& will be sent to Hui Lai Garden.& will be sent to Hui Lai Garden.\\
&&&&I live in Shanghai.\\
\midrule
\multirow{3}{*}{175-185} &\multirow{3}{*}{1}&...... but the display effect is good&...... but the screen display effect&...... but the screen display effect\\
&&the accessories are complete.&is good, and the accessories&is good, and the accessories\\
&&&are complete&are complete.\\
\midrule
\multirow{2}{*}{197-205} &\multirow{2}{*}{1}&...... But no delivery person came& ...... But there is no rider to& ...... But there is no rider to\\
&& to the store to pick it up.& take away the food& take away the food.\\
\midrule
\multirow{6}{*}{228-234} &\multirow{6}{*}{3}& ...... Nourish the heart and calm the & ...... It is also suitable for patients& ...... It is also suitable for patients\\
&& nerves, treat palpitations, insomnia, &with poor blood and insomnia.&with poor blood, insomnia and\\
&&stomach deficiency, forgetfulness,&&other symptoms.\\
&&vexation and thirstiness, nourish and &&\\
&&replenish deficiency, and prevent sper-&&\\
&&matorrhea and premature ejaculation.&&\\
\midrule
\multirow{5}{*}{284-295} &\multirow{5}{*}{3}&...... Virgin olive oil, also known&...... Virgin olive oil is the olive&...... Virgin olive oil is the olive\\
&&as natural olive oil, is oil obtained& oil directly obtained by mechanical& oil directly obtained by mechanical\\
&&from fresh olive fruit after removing& cold pressing from fresh olives.& cold pressing from fresh olives,\\
&&foreign matters by mechanical cold &&and refined olive oil is the olive oil\\
&&pressing and filtering.&& obtained by refining virgin olive oil.\\
\midrule
\multirow{4}{*}{443-451} &\multirow{4}{*}{3}&...... Pang Yaotian (the one in the & ...... Xinhua Xinhua News Agency,& ...... Xinhua Xinhua News Agency,\\
&&middle), a player of Beijing Dream&December 11.&December 11. Beijing Dreaming\\
&&ING team, was shooting in the &&Team player Pang Yitian (C)\\
&&game on December 11. &&shoots during the game.\\
\midrule
\multirow{3}{*}{473-481} &\multirow{3}{*}{2}&...... it can also let you play CSgo&...... CSgo is also easy to win,&...... CSgo is also easy to win.\\
&&without any trouble and you can&most three A games can be played&Most three-A games can be played.\\
&&play most Three A games on it.&&\\
\midrule
\multirow{8}{*}{507-508} &\multirow{8}{*}{3}&Therefore, ZTCHINA, which ......,&Therefore, ......, and won the&Therefore, ......, and has won the\\
&&and won the honorary title of “China & honorary title of "Chinese&honorary title of "Chinese\\
&&Enterprise R\&D and Management&enterprise research and&enterprise research and development\\
&&Talents Training Demonstration Base”,&development management talent& management talent training demons-\\
&&presents the online course R\&D&training demonstration base".& tration base". With the online course\\
&&Project Manager Special Training&& "Research and Development Project\\
&&Camp.&& Manager Special Training Camp" \\
&&&&led by Zhongtian Huaxia Consulting.\\
\midrule
\multirow{2}{*}{552-557} &\multirow{2}{*}{1}&...... you must refund the deposit&...... It must be returned &...... It must be returned\\
&&and cancel the order. OK.&and cancel the order. Good&and cancel the order. Good.\\
\midrule
\multirow{3}{*}{558-563} &\multirow{3}{*}{3}&...... when my myvi just reaches & ...... That is to say, myvi can only& ...... That is to say, myvi can only\\
&&30km/h, it is already at 100km/h,& reach 30 km/h when others reach& reach 30 km/h when others reach\\
&& hum.& 100 km/h.& 100 km/h. Hmm.\\
\bottomrule 
\end{tabular}
\end{small}
}
\end{center}
\caption{Translation differences after enhancing the EOS penalty (WMT22 Zh-En test set, Document 1-100).}
\label{table:diffdocall}
\end{table}

\newpage
\section{Effect of Hyperparameters}
There are hyperparameters in the proposed EOS penalty, such as $\tau=1$ in Equation \ref{eq:cond} and $\beta=0.4$ in Equation \ref{eq:5}. The hyperparameter $\tau=1$ controls the range at which the EOS penalty takes effect, and the hyperparameter $\beta$ represents the weight of the EOS penalty. A larger $\tau$ or $\beta$ can fix more under-translations but also introduce more other modifications.

The hyperparameters were chosen based on extensive experiments, and finally the choosen hyperparameters were found to exhibit stable performance across different datasets. Overall, our method exhibits relatively low sensitivity to these hyperparameters. Even with significant adjustments to the hyperparameters (e.g., increasing $\tau$ to 2 or decreasing $\beta$ to 0.2), the impact on translations remains small. To illustrate it, we generate translations of the WMT 22 Zh-En test set under different hyperparameter settings. The BLEU scores are displayed in Table \ref{table:hyp1}, showing that the impact of hyperparameters on overall translation quality is relatively minor. The proportions of under-translation and over-translation on the first 300 sentences of WMT22 Zh-En test set are displayed in Table \ref{table:hyp2}, indicating that tweaking the hyperparameters can slightly affect the rates of over-translation and under-translation.

\begin{table}[h]
\begin{center}
\begin{small}
\begin{tabular}{lcccc}
\toprule
\bf System & w/o EOS penalty & $\tau=1, \beta=0.4$& $\tau=2, \beta=0.4$&$\tau=1, \beta=0.2$ \\
\midrule
\bf BLEU & 23.34 & 23.47 & 23.48 & 23.45 \\
\bottomrule
\end{tabular}
\end{small}
\end{center}
\caption{BLEU scores on WMT22 Zh-En test sets.}
\label{table:hyp1}
\vspace{-1em}
\end{table}

\begin{table}[h]
\begin{center}
\begin{small}
\begin{tabular}{lcccc}
\toprule
\bf System & w/o EOS penalty & $\tau=1, \beta=0.4$& $\tau=2, \beta=0.4$&$\tau=1, \beta=0.2$ \\
\midrule
\bf Under & 9.7\% & 6.7\% & 6.7\% & 7.0\% \\
\bf Over & 2.3\% & 2.3\% & 2.6\% & 2.3\% \\
\bottomrule
\end{tabular}
\end{small}
\end{center}
\caption{The proportions of under-translation and over-translation on the subset of WMT22 Zh-En test set.}
\label{table:hyp2}
\end{table}

Overall, our method exhibits relatively low sensitivity to hyperparameters, affecting only a small number of translations. For instance, Table \ref{table:beta} illustrates the impact of reducing $\beta$ to 0.2 on the translations of first 300 sentences of WMT22 Zh-En test set, resulting in changes in just two translations.

\begin{table}[h]
\begin{center}
\begin{small}
\begin{tabular}{cccc}
\toprule
\bf Line  & \bf Reference & $\tau=1, \beta=0.2$ & $\tau=1, \beta=0.4$ \\
\midrule
\multirow{2}{*}{2} & So that such a thing won’t& So as not to happen again & So as not to happen again.\\
&happen again.&&\\
\midrule
\multirow{5}{*}{133} &Inspection shaft concrete module&Inspection of the production and&Inspection of the production and\\
&production and processing &processing equipment of the&processing equipment of the\\
&equipment, the assembly site for&inspection well concrete block&inspection shaft concrete module\\
&Zhengzhou shaft wall brick&module:&Zhengzhou inspection shaft brick \\
&module equipment&&module production line:\\
\bottomrule 
\end{tabular}
\end{small}
\end{center}
\caption{Translation differences between hyperparameter settings ($\tau=1, \beta=0.2$) and ($\tau=1, \beta=0.4$) on the subset of WMT22 Zh-En test set.}
\label{table:beta}
\end{table}

\section{Encoder-Decoder Transformer}
In this study, our experiments are mainly conducted on the fine-tuned LLaMA2 model, due to its superior translation capabilities and support for document-level translation, aligning more closely with real-world applications. Our method is also applicable to the traditional encoder-decoder Transformer architecture. Specifically, when applied to the Transformer-big model \citep{vaswani2017attention} trained on the WMT22 Zh-En dataset, the application of EOS penalty results in an improvement of the BLEU score from 15.25 to 15.36. After applying the EOS penalty, we observe changes in the translations within the first 300 sentences of the WMT22 Zh-En test set, with a total of 18 instances of modification. Among these, 4 instances of under-translations are corrected. 
\section{Pseudo Code for EOS Penalty}
\lstset{language=Python,
        basicstyle=\ttfamily\small,
        keywordstyle=\color{blue},
        stringstyle=\color{red},
        commentstyle=\color{red},
        numbers=left,
        numberstyle=\tiny\color{gray},
        stepnumber=1,
        numbersep=10pt,
        backgroundcolor=\color{white},
        showspaces=false,
        showstringspaces=false,
        showtabs=false,
        frame=single,
        rulecolor=\color{black},
        tabsize=2,
        captionpos=b,
        breaklines=true,
        breakatwhitespace=false,
        title=\lstname,
        escapeinside={\%*}{*)},
        morekeywords={*,...}
}
\begin{lstlisting}
# Initialization
beam_scores = initialize_zeros(batch_size, num_beams)
input_ids = initialize_with_start_token_ids()
cur_len = initial_sequence_length
tau = 1
beta = 0.4
L = 20

# Generation loop
while not all_beams_ended:
    # Generate the probability distribution for the next token
    next_token_logits = model(input_ids)
    next_token_scores = F.log_softmax(next_token_logits, dim=-1).view(batch_size, num_beams, vocab_size)
    
    # Detect under-translation by EOS probability
    eos_score = next_token_scores[:, :, EOS_token_id]
    next_highest_score = torch.topk(next_token_scores, 2, dim = -1)[0][:, :, 1]
    eos_detected = (eos_score - next_highest_score) < tau

    # Generate top beam candidates
    next_token_scores = next_token_scores + beam_scores[:, None].expand_as(next_token_scores)
    next_token_scores = next_token_scores.view(batch_size, num_beams * vocab_size)
    next_token_scores, next_tokens = torch.topk(
        next_token_scores, 2 * num_beams, dim=1, largest=True, sorted=True
    ) # (batch_size, 2 * num_beams)

    # Apply EOS penalty on beam candidates
    for batch_idx in range(batch_size):
        for token_idx in range(num_beams * 2):
            next_token = next_tokens[batch_idx, token_idx]
            # if the candidate is finalized, check and apply EOS penalty
            if next_tokens % vocab_size == EOS_token_id:
                beam_idx = next_token // vocab_size
                if eos_detected[batch_idx][beam_idx]:
                    eos_penalty = min(cur_len - initial_len, L) * beta * eos_score[idx][beam_idx]
                    next_token_scores[batch_idx][token_idx] += eos_penalty

    # Update beam scores based on next_token_scores
    update_beam_scores(next_token_scores)
    
    # Select the next tokens and their corresponding beams
    next_tokens, next_beam_indices = select_topk(next_token_scores)
    update_input_ids_and_beam_indices(next_tokens, next_beam_indices)
    
    # Update current length
    cur_len += 1

# Finalization
sequences = finalize_sequences_based_on_beam_scorer()
return sequences

\end{lstlisting}

\end{document}